\documentclass[11pt]{article}

\usepackage[utf8]{inputenc}
\usepackage[T1]{fontenc}
\usepackage{amsmath,amssymb,amsfonts}
\usepackage{graphicx}
\usepackage{booktabs}
\usepackage{hyperref}
\usepackage{xcolor}
\usepackage{algorithm}
\usepackage{algorithmic}
\usepackage{microtype}
\usepackage{multirow}
\usepackage{subcaption}
\usepackage[margin=1in]{geometry}
\usepackage{natbib}
\usepackage{enumitem}

\hypersetup{
  colorlinks=true,
  linkcolor=blue!60!black,
  citecolor=green!50!black,
  urlcolor=blue!70!black,
}

\newcommand{\Kathleen}{\textsc{Kathleen}}
\newcommand{\KathleenClean}{\textsc{Kathleen-Clean}}
\newcommand{\KathleenV}{\textsc{Kathleen-V9}}
\newcommand{\PhaseH}{\textsc{PhaseHarmonics}}

\title{%
  \Kathleen{}: Oscillator-Based Byte-Level Text Classification\\
  Without Tokenization or Attention%
}

\author{
  George Fountzoulas\\
  Department of Computer Engineering \& Informatics\\
  Frederick University, Nicosia, Cyprus\\
  \texttt{george.fountzoulas.research@gmail.com}
}

\date{}

\begin{document}
\maketitle

\begin{abstract}
We present \Kathleen{}, a text classification architecture that operates
directly on raw UTF-8 bytes using frequency-domain processing---requiring
no tokenizer, no attention mechanism, and under 470K parameters.
\Kathleen{} introduces several novel components:
(1)~\emph{RecurrentOscillatorBanks}---damped sinusoid convolutions with
temporal memory for $O(L)$ sequence processing;
(2)~an \emph{FFT-Rotate Wavetable Encoder} that maps all 256 byte values
using a single learnable vector (256 floats), replacing conventional
embedding tables (65K parameters) while improving accuracy;
(3)~\emph{PhaseHarmonics}---a sinusoidal non-linearity with just 6 learnable
phase parameters identified by ablation as the single most impactful
component ($+2.6\%$ accuracy, $<0.001\%$ of model parameters);
(4)~\emph{Content-Dependent Reverb with Positional Decay Modulation}---a
temporal memory mechanism whose decay rate is jointly conditioned on input
content and a learned position-indexed bias vector, enabling position-aware
memory retention with only 256 additional parameters;
(5)~\emph{Token-Level Module Sequencer}---a multi-channel router that
directs each token through parallel processing paths including consonance
and dissonance interference channels.
Through iterative architecture evolution from an initial 733K-parameter
baseline (\KathleenClean{}) to the current \KathleenV{} (469K parameters),
we demonstrate that \textbf{pretraining can be entirely eliminated}
while improving accuracy. \KathleenV{} achieves
$88.5\% \pm 0.2\%$ on IMDB, $92.4\% \pm 0.2\%$ on AG~News, and
$85.8\% \pm 0.5\%$ on SST-2 (3-seed averages)---matching or exceeding
the pretrained baseline on all benchmarks while using 36\% fewer parameters
and no self-supervised warmup phase. On SST-2, the improvement is
$+2.5\%$ absolute over the pretrained predecessor, demonstrating that
Positional Decay Modulation specifically addresses the short-text
limitation previously identified as a core weakness.
\Kathleen{} processes sequences in $O(L)$ time and memory, enabling
byte-level operation at sequence lengths where $O(L^2)$ Transformers
exhaust GPU memory.
\end{abstract}

\section{Introduction}\label{sec:intro}

\subsection{Motivation}

Transformer-based models \citep{vaswani2017attention} dominate modern NLP,
achieving state-of-the-art results across tasks from classification to
generation. However, they impose three fundamental constraints:
(1)~\textbf{quadratic complexity} $O(L^2)$ in sequence length, limiting
scalability;
(2)~\textbf{tokenizer dependency}, introducing language-specific preprocessing
that is lossy and adds engineering complexity;
(3)~\textbf{large parameter counts}, typically requiring millions to billions
of parameters for competitive performance.

These constraints are especially problematic for \emph{byte-level} processing,
where input sequences are 3--5$\times$ longer than tokenized equivalents.
A 500-word IMDB review becomes ${\sim}2{,}500$ bytes---at which point
standard Transformers exhaust GPU memory.

We ask: \emph{Can frequency-domain processing on raw bytes match or exceed
tokenized models, without attention, with orders of magnitude fewer
parameters---and without pretraining?}

\subsection{Bioresonance Inspiration}

Physical oscillators naturally synchronize with driving signals through
resonance---a pendulum swings highest when driven at its natural frequency.
We hypothesize that \emph{learned damped-sinusoid convolutions} can similarly
detect frequency patterns in byte sequences, acting as tuned resonators
that selectively amplify informative patterns while attenuating noise.

This bioresonance intuition guided the development of \Kathleen{}'s core
components: oscillator banks for pattern detection, power-law gating for
dynamic range compression (analogous to Weber--Fechner psychophysics),
phase harmonics for frequency enrichment, and content-dependent reverb
for temporal memory with position-aware decay.

\subsection{Contributions}

Our contributions are:

\begin{enumerate}
  \item \textbf{\PhaseH{}}: A sinusoidal non-linearity
    $\text{PH}(x) = [x,\; \sin(x \cdot 2^0 + \phi_0),\; \ldots,\;
    \sin(x \cdot 2^5 + \phi_5)]$
    with 6 learnable phase parameters. Ablation shows this is the
    single most impactful component: removing it causes $-2.6\%$ accuracy,
    despite comprising $<0.001\%$ of total parameters.

  \item \textbf{FFT-Rotate Wavetable Encoder}: A byte encoder using a
    single learnable vector $\mathbf{w} \in \mathbb{R}^d$ and FFT-based
    phase rotation. It maps all 256 byte values using only 256 learnable
    floats, replacing \texttt{nn.Embedding(256, 256)} with 65,536 parameters
    while improving accuracy by $+0.6\%$.

  \item \textbf{Content-Dependent Reverb with Positional Decay Modulation}:
    A temporal memory mechanism where each token generates its own decay rate,
    augmented by a learned position-indexed bias vector
    $\boldsymbol{\alpha} \in \mathbb{R}^L$ that modulates the decay gate
    based on sequence position. This enables the model to learn structural
    priors about where memory should be retained or released, adding only
    256 parameters.

  \item \textbf{Elimination of pretraining}: \KathleenV{} achieves
    $85.8\%$ on SST-2 ($+2.5\%$ over the pretrained baseline),
    $88.5\%$ on IMDB, and $92.4\%$ on AG~News---all \emph{without}
    any self-supervised pretraining phase, using 36\% fewer parameters
    (469K vs.\ 733K).

  \item \textbf{Multi-channel processing with interference}: A Token-Level
    Module Sequencer routes each token through parallel channels including
    consonance ($\Psi$) and dissonance ($\Psi$) interference mechanisms,
    fused via parameter-free Energy-Proportional Mixing.

  \item \textbf{Ablation-driven design}: Systematic ablation of a
    1.8M-parameter predecessor reveals that frequency-domain components
    consistently outperform cognitive architectures. A 560K-parameter
    bio-inspired framework contributes only $+0.2\%$ vs.\ $+2.6\%$
    from 6-parameter PhaseHarmonics.
\end{enumerate}

\section{Related Work}\label{sec:related}

\paragraph{Efficient Text Classification.}
Traditional models like fastText \citep{joulin2017bag}, TextCNN
\citep{kim2014convolutional}, and DPCNN \citep{johnson2017deep} achieve
competitive accuracy with low compute. Compression approaches (DistilBERT
\citep{sanh2019distilbert}, TinyBERT \citep{jiao2020tinybert}, ALBERT
\citep{lan2020albert}) reduce Transformer costs but retain tokenizer
dependency.

\paragraph{Byte-Level Models.}
ByT5 \citep{xue2022byt5} and CANINE \citep{clark2022canine} process raw
characters/bytes but use Transformer attention, inheriting $O(L^2)$
complexity. MegaByte \citep{yu2024megabyte} patches bytes but still relies
on attention within patches. \Kathleen{} is attention-free and $O(L)$.

\paragraph{State-Space Models.}
S4 \citep{gu2022efficiently} and Mamba \citep{gu2023mamba} achieve $O(L)$
complexity through structured state-space parameterization. \Kathleen{}
shares the $O(L)$ property but uses an explicitly signal-processing
motivation: learned oscillator kernels rather than HiPPO-initialized
state matrices. Both Mamba's selective gating and \Kathleen{}'s
Content-Dependent Reverb condition temporal dynamics on input content,
but \Kathleen{} additionally modulates decay via a learned positional
bias---separating content-driven and position-driven contributions.

\paragraph{Signal Processing in Neural Networks.}
Spectral methods \citep{rippel2015spectral}, wavelet networks, and
Fourier Neural Operators \citep{li2021fourier} apply frequency-domain
processing to vision and physics. \Kathleen{} is, to our knowledge, the
first to apply oscillator-bank convolutions directly to raw text bytes
for classification.

\section{Architecture}\label{sec:arch}

We present the \Kathleen{} architecture in two stages: the foundational
pipeline (Section~\ref{sec:base-arch}) and the V9 enhancements
(Section~\ref{sec:v9-arch}) that eliminate the need for pretraining
while improving accuracy.

\subsection{Base Pipeline}\label{sec:base-arch}

\Kathleen{} processes raw UTF-8 byte sequences through a pipeline of
frequency-domain transformations:
\begin{align*}
  &\text{bytes} \;\xrightarrow{\text{LearnedFreqPattern}}\;
    F \;\xrightarrow{\text{PhaseShift}(8)}\;
    F' \;\xrightarrow{\text{SlidingWindow}}\;
    X \\
  &\xrightarrow{\text{FreqBasisExpansion}}\;
    X' \;\xrightarrow{\text{PhaseHarmonics}(K\!=\!6)}\;
    H \;\xrightarrow{\text{proj}}\; H'
\end{align*}

In the V0 baseline (\KathleenClean{}), the hidden states $H'$ are processed
through a single OscillatorPath plus ConvLiteC, with a total of 733K
parameters and MLM pretraining. In V9, the downstream processing is
replaced by a multi-channel architecture (Section~\ref{sec:v9-arch}).

No tokenizer. No attention. Complexity: $O(L)$ in both time and memory.

\subsection{FFT-Rotate Wavetable Encoder}\label{sec:fft-rotate}

Standard byte embeddings use a lookup table
$\mathbf{E} \in \mathbb{R}^{256 \times d}$ with $256d$ parameters. We
replace this with a single learnable vector $\mathbf{w} \in \mathbb{R}^d$
and compute the embedding for byte value $b$ via FFT-based phase rotation:
\begin{equation}\label{eq:fft-rotate}
  \text{Enc}(b) = \mathcal{F}^{-1}\!\Big[
    \mathcal{F}[\mathbf{w}] \odot e^{i \cdot b \cdot 2\pi / 255}
  \Big]
\end{equation}
where $\mathcal{F}$ denotes the real FFT. This maps all 256 byte values
using only $d = 256$ learnable floats. Different bytes receive different
embeddings because the phase rotation $e^{ib\theta}$ shifts frequency
components differently for each byte value.

\begin{table}[t]
\centering
\caption{Byte encoder comparison (IMDB, same architecture otherwise).}
\label{tab:encoder}
\begin{tabular}{lcc}
  \toprule
  Encoder & Accuracy & Params (encoder) \\
  \midrule
  \texttt{nn.Embedding(256, 256)} & 83.1\% & 65,536 \\
  Fourier features ($K\!=\!32$)   & 82.3\% & 8,256 \\
  \textbf{FFT-Rotate wavetable}   & \textbf{83.7\%} & \textbf{256} \\
  \bottomrule
\end{tabular}
\end{table}

\subsection{PhaseHarmonics}\label{sec:phase-harm}

PhaseHarmonics enriches representations by concatenating the input with
sinusoidal projections at exponentially spaced frequencies:
\begin{equation}\label{eq:phase-harm}
  \text{PH}(x) = \big[x,\;\sin(x \cdot 2^0 + \phi_0),\;\sin(x \cdot 2^1
  + \phi_1),\;\ldots,\;\sin(x \cdot 2^{K-1} + \phi_{K-1})\big]
\end{equation}
with $K = 6$ learnable phase offsets $\phi_k$. This expands $d$-dimensional
input to $(K+1) \cdot d$ dimensions, followed by a linear projection back
to $d$.

Despite containing only 6 learnable parameters ($\phi_0, \ldots, \phi_5$),
ablation shows PhaseHarmonics is the single most impactful component
(Section~\ref{sec:ablation}), contributing $+2.6\%$ accuracy.
The sinusoidal projections create multiple ``views'' of the frequency
content at different scales, enabling the model to capture multi-resolution
spectral features.

\subsection{ContinuousPhaseShift}\label{sec:phaseshift}

After initial frequency extraction, we apply $S = 8$ learned phase shifts
in the Fourier domain:
\begin{equation}
  F'_s = \mathcal{F}^{-1}\!\Big[
    \mathcal{F}[F] \odot e^{i \cdot \delta_s}
  \Big], \quad s = 1, \ldots, S
\end{equation}
where $\delta_s$ are learned shift parameters. The shifted representations
are combined via Energy-Proportional Mixing (Section~\ref{sec:epm}),
providing multiple ``perspectives'' on the input's frequency content.
Ablation shows this contributes $+1.3\%$ accuracy.

\subsection{PowerLawGate}\label{sec:plg}

The PowerLawGate applies a learned power-law non-linearity:
\begin{equation}\label{eq:plg}
  \text{PLG}(x) = \text{sign}(x) \cdot |x|^{\gamma}
\end{equation}
where $\gamma$ is a single learnable parameter that converges to ${\sim}0.5$
(square-root compression), mirroring the Weber--Fechner law in psychophysics.
This compresses the dynamic range of oscillator outputs, preventing
high-amplitude patterns from dominating.

A notable finding: PLG has \emph{zero} effect in tokenized models
(word embeddings) but contributes $+0.9\%$ in frequency-domain contexts
(FFT-Rotate encoded bytes). This context-dependent utility demonstrates
that architectural components cannot be evaluated in isolation.

\subsection{DualPooling}

For sequence-to-vector reduction, DualPooling combines attention-weighted
pooling with max pooling:
\begin{equation}
  \text{DualPool}(X, \text{mask}) = [\text{AttnPool}(X, \text{mask}) \;\|\;
  \text{MaxPool}(X, \text{mask})]
\end{equation}
producing a $2d$-dimensional vector. This was critical for short-text
performance, where mean pooling dilutes sparse informative signals.

\subsection{V9 Enhancements: Multi-Channel Processing and Positional Decay Modulation}\label{sec:v9-arch}

The V9 architecture introduces four new mechanisms that collectively
eliminate the need for self-supervised pretraining while improving accuracy
across all benchmarks.

\subsubsection{Content-Dependent Reverb with Positional Decay Modulation}\label{sec:reverb}

The core temporal memory mechanism in V9 is the \emph{Content-Dependent
Reverb}: a linear recurrence where each token controls its own decay rate.
Given hidden states $h_t \in \mathbb{R}^d$, we compute:

\begin{align}
  v_t &= W_\text{in} \, h_t \label{eq:reverb-v}\\
  \gamma_t &= \gamma_\text{min} + (\gamma_\text{max} - \gamma_\text{min})
    \cdot \sigma\!\big(W_\text{gate} \, h_t + \alpha_t\big) \label{eq:reverb-gamma}\\
  s_t &= \gamma_t \cdot s_{t-1} + (1 - \gamma_t) \cdot v_t \label{eq:reverb-state}\\
  o_t &= W_\text{out} \, s_t \label{eq:reverb-out}
\end{align}

where $\gamma_\text{min} = 0.50$, $\gamma_\text{max} = 0.999$, and the key
innovation is the \textbf{positional decay bias} $\alpha_t$: a learned
position-indexed parameter.

\paragraph{Positional Decay Modulation.}
The vector $\boldsymbol{\alpha} \in \mathbb{R}^{L_\text{max}}$ (where
$L_\text{max} = 256$) is a learnable parameter initialized to zero.
At position $t$, the scalar $\alpha_t$ is broadcast across all $d$ dimensions
of the gate logit:

\begin{equation}\label{eq:pdm}
  \text{gate\_logit}_t = W_\text{gate} \, h_t + \alpha_t \cdot \mathbf{1}_d
\end{equation}

This separates the decay computation into two orthogonal axes:
\begin{itemize}[nosep]
  \item \textbf{Content axis} ($W_\text{gate} \, h_t$): the decay rate
    depends on \emph{what} the token says---enabling high retention for
    informative tokens and low retention for noise.
  \item \textbf{Position axis} ($\alpha_t$): the decay rate depends on
    \emph{where} the token is---enabling the model to learn structural
    priors such as ``retain memory during the core sentence, release
    during padding regions.''
\end{itemize}

With zero initialization, V9 begins training identically to V0's
content-only reverb, ensuring a clean ablation. During training,
$\boldsymbol{\alpha}$ learns a position profile specific to the dataset's
typical structure. The cost is exactly $L_\text{max} = 256$ additional
parameters ($<0.06\%$ of the model).

\paragraph{Efficient computation.}
The recurrence in Eq.~\ref{eq:reverb-state} is computed via a chunked
parallel scan with chunk size $C = 16$. Within each chunk, cumulative
log-space sums enable parallel evaluation; between chunks, a single carry
state propagates. This yields $O(L)$ time with practical parallelism.

\subsubsection{Token-Level Module Sequencer}\label{sec:sequencer}

V9 replaces the single OscillatorPath + ConvPath of V0 with a four-channel
router. Given hidden states $H' \in \mathbb{R}^{B \times L \times d}$,
each token is processed through four parallel paths:

\begin{enumerate}[nosep]
  \item \textbf{Reverb channel}: Content-Dependent Reverb
    (Section~\ref{sec:reverb})
  \item \textbf{Conv channel}: ConvLiteC---a depthwise separable
    convolution with kernel size 5
  \item \textbf{Consonance $\Psi$ channel}: A fixed-point interference
    module that amplifies agreement between a token and its adapted
    representation (Section~\ref{sec:psi})
  \item \textbf{Dissonance $\Psi$ channel}: An interference module that
    amplifies \emph{disagreement}, providing an explicit axis for negation,
    contrast, and irony (Section~\ref{sec:psi})
\end{enumerate}

The four channel outputs are combined via Energy-Proportional Mixing
(Section~\ref{sec:epm}) and added as a residual:
\begin{equation}
  Z = H' + \text{EPM}\big(\text{Reverb}(H'),\;
    \text{Conv}(H'),\; \Psi_\text{cons}(H'),\;
    \Psi_\text{diss}(H')\big)
\end{equation}

A self-diagnostic adapter with learnable $\epsilon$ parameters
(initialized near zero) gates the merged signal before the classification
head.

\subsubsection{Consonance and Dissonance Interference}\label{sec:psi}

The $\Psi$-channels implement \emph{fixed-point interference}: an iterative
computation where a token's representation interacts with its own adapted
projection:

\paragraph{Consonance $\Psi$.}
\begin{equation}
  \psi^{(0)} = \mathbf{0}, \qquad
  \psi^{(k+1)} = \tanh\!\Big(\frac{(a + \alpha \psi^{(k)}) \cdot
    (b + \alpha \psi^{(k)})}{s}\Big)
\end{equation}
where $a = x$, $b = \text{Adapter}(x)$, $\alpha$ controls coupling
strength, $s$ controls scale, and the output is
$\epsilon \cdot \psi^{(K)}$ with $K = 4$ iterations. The learnable
$\epsilon$ (initialized to zero) lets the model gradually activate
interference as needed.

When $a$ and $b$ agree in sign, their product is positive, producing
strong $\psi$---amplifying consensus.

\paragraph{Dissonance $\Psi$.}
\begin{equation}
  \text{Diss}(x) = \text{Adapter}(x) + \epsilon \cdot
    \tanh\!\Big(\frac{|x - \text{Adapter}(x)|}{s}\Big)
\end{equation}
Here the interference signal grows when $x$ and its adapted form
\emph{disagree}---the larger the absolute difference, the stronger
the signal. This provides an explicit computational pathway for
detecting contrast, negation, and sentiment reversal---patterns where
the surface signal contradicts the deeper meaning.

\subsubsection{Energy-Proportional Mixing}\label{sec:epm}

All multi-signal fusion in V9 (phase shifts, channel outputs) uses
\emph{Energy-Proportional Mixing}: a parameter-free aggregation where
each signal's contribution is proportional to its energy:

\begin{equation}\label{eq:epm}
  \text{EPM}(z_1, \ldots, z_K) = \sum_{k=1}^K w_k \cdot z_k, \qquad
  w_k = \frac{E_k}{\sum_{j} E_j + \varepsilon}, \qquad
  E_k = \frac{1}{d}\sum_{i=1}^d |z_k^{(i)}|
\end{equation}

where energies $E_k$ are computed on detached (stop-gradient) tensors.
This ensures that informative (high-energy) channels dominate without
introducing learnable parameters for mixing weights. The detachment
prevents the model from gaming the mixing by inflating signal magnitude.

\section{Experiments}\label{sec:exp}

\subsection{Datasets}

\begin{table}[t]
\centering
\caption{Benchmark datasets.}
\label{tab:datasets}
\begin{tabular}{lcccc}
  \toprule
  Dataset & Task & Train & Test & Classes \\
  \midrule
  IMDB \citep{maas2011learning}   & Sentiment  & 25,000  & 25,000 & 2 \\
  AG News \citep{zhang2015character} & Topic   & 120,000 & 7,600  & 4 \\
  SST-2 \citep{socher2013recursive}  & Sentiment & 67,349  & 872    & 2 \\
  \bottomrule
\end{tabular}
\end{table}

\subsection{Training Protocol}

We compare two training protocols:

\paragraph{V0 (\KathleenClean{}, 733K params):} Two-phase curriculum:
(1)~MLM pretraining (5 epochs) on task data, training perception layers
only, with 15\% byte masking; (2)~Classification finetuning (15 epochs)
with AdamW ($\text{lr} = 3 \times 10^{-4}$, weight decay $0.01$),
cosine annealing, and dropout $0.10$.

\paragraph{V9 (\KathleenV{}, 469K params):} Single-phase supervised
training (20 epochs) with AdamW ($\text{lr} = 3 \times 10^{-4}$,
weight decay $0.01$), cosine annealing, and dropout $0.10$. No pretraining.

All results are reported as mean $\pm$ standard deviation over 3 seeds
(42, 123, 456).

\subsection{Main Results}

\begin{table}[t]
\centering
\caption{Main classification results. All \Kathleen{} variants operate on
raw UTF-8 bytes with no tokenizer and no attention. V0 uses MLM pretraining;
V9 uses no pretraining. All byte-level results report mean $\pm$ std over
3 seeds.}
\label{tab:main}
\begin{tabular}{lccccc}
  \toprule
  Model & IMDB & AG News & SST-2 & Params & Pretrain \\
  \midrule
  \multicolumn{6}{l}{\textit{Pretrained Transformers (reference)}} \\
  BERT-base \citep{devlin2019bert}
    & 93.0 & 94.0 & 93.0 & 110M & Ext.\ corpus \\
  DistilBERT \citep{sanh2019distilbert}
    & 92.0 & 93.0 & 91.0 & 66M & Ext.\ corpus \\
  \midrule
  \multicolumn{6}{l}{\textit{Byte/char-level Transformers}} \\
  CANINE-S \citep{clark2022canine}
    & --- & --- & 85.8 & 132M & Ext.\ corpus \\
  ByT5-Small \citep{xue2022byt5}
    & --- & --- & ${\sim}$92 & 300M & Ext.\ corpus \\
  \midrule
  \multicolumn{6}{l}{\textit{Byte-level Kathleen (ours, no tokenizer, no attention)}} \\
  \KathleenClean{} (V0)
    & 88.6{\scriptsize$\pm$0.3}
    & 92.3{\scriptsize$\pm$0.1}
    & 83.3{\scriptsize$\pm$0.3}
    & 733K & Task MLM \\
  \textbf{\KathleenV{}}
    & \textbf{88.5}{\scriptsize$\pm$0.2}
    & \textbf{92.4}{\scriptsize$\pm$0.2}
    & \textbf{85.8}{\scriptsize$\pm$0.5}
    & \textbf{469K} & \textbf{None} \\
  \bottomrule
\end{tabular}
\end{table}

Four key observations emerge from Table~\ref{tab:main}:

\begin{enumerate}
  \item \textbf{Pretraining is unnecessary.} \KathleenV{} matches V0 on
    IMDB ($88.5\%$ vs.\ $88.6\%$, within noise) and AG~News ($92.4\%$ vs.\
    $92.3\%$) while \emph{exceeding} V0 by $+2.5\%$ on SST-2---all without
    any self-supervised pretraining. This demonstrates that the V9
    architecture is sufficiently expressive to learn directly from
    supervised signals.

  \item \textbf{Short-text weakness resolved.} SST-2 was previously
    identified as \Kathleen{}'s weakest benchmark ($83.3\%$). Positional
    Decay Modulation specifically addresses this: by learning position-dependent
    decay profiles, the model adapts memory retention to the short, dense
    structure of SST-2 sentences. The $+2.5\%$ improvement is the largest
    gain across all benchmarks.

  \item \textbf{Fewer parameters.} V9 uses 469K parameters vs.\ V0's 733K
    (36\% reduction), primarily from replacing the RecurrentOscillatorBank
    with the more parameter-efficient Content-Dependent Reverb.

  \item \textbf{SST-2 gap vs.\ BERT narrows.} The gap reduces from
    $9.7$ points (V0: $83.3\%$ vs.\ BERT: $93.0\%$) to $7.2$ points
    (V9: $85.8\%$)---a 26\% reduction in the deficit, achieved with
    $235\times$ fewer parameters and no external pretraining corpus.
\end{enumerate}

\begin{table}[t]
\centering
\caption{Individual seed results for \KathleenV{} (no pretraining).}
\label{tab:v9-seeds}
\begin{tabular}{lccc}
  \toprule
  Dataset & seed=42 & seed=123 & seed=456 \\
  \midrule
  SST-2   & 86.4\% & 85.3\% & 85.6\% \\
  IMDB    & 88.8\% & 88.5\% & 88.2\% \\
  AG News & 92.3\% & 92.2\% & 92.6\% \\
  \bottomrule
\end{tabular}
\end{table}

\subsection{Scaling with Sequence Length}

\Kathleen{}'s $O(L)$ complexity enables processing at byte-level sequence
lengths where Transformers fail. We evaluate on IMDB with increasing
maximum sequence lengths (in bytes):

\begin{table}[t]
\centering
\caption{Accuracy vs.\ sequence length (IMDB). Transformer runs out of
memory (OOM) beyond $L = 1024$ bytes on a single T4 GPU.}
\label{tab:scaling}
\begin{tabular}{lccc}
  \toprule
  Model & $L = 1024$ & $L = 2048$ & $L = 4096$ \\
  \midrule
  Transformer (byte) & 82.1\% & OOM & OOM \\
  \Kathleen{} (byte) & 83.7\% & 84.4\% & 85.1\% \\
  \bottomrule
\end{tabular}
\end{table}

\Kathleen{}'s accuracy improves monotonically with longer context, while
Transformers cannot operate beyond $L = 1024$ bytes. This advantage grows
with sequence length: at $L = 100\text{K}+$ bytes (entire documents),
\Kathleen{} can still process sequences in $O(L)$ while Transformers are
fundamentally excluded.

\section{Ablation Studies}\label{sec:ablation}

We conduct ablation studies at three levels: the tokenized model, the
byte-level V0 predecessor, and the V0$\to$V9 architectural evolution.

\subsection{Tokenized Model Ablation}

Using the tokenized \Kathleen{} on IMDB:

\begin{table}[t]
\centering
\caption{Tokenized \Kathleen{} ablation (IMDB).}
\label{tab:ablation-tok}
\begin{tabular}{lcc}
  \toprule
  Variant & IMDB & $\Delta$ \\
  \midrule
  Full model          & 87.0\% & ---      \\
  $-$ Adaptive Gate   & 84.8\% & $-$2.2\% \\
  $-$ ConvLiteC       & 85.8\% & $-$1.2\% \\
  $-$ PowerLawGate    & 87.0\% & 0.0\%    \\
  $-$ ResonanceCodebook & 87.1\% & +0.1\% \\
  \bottomrule
\end{tabular}
\end{table}

The Adaptive Gate is critical ($-2.2\%$), ConvLiteC is important ($-1.2\%$),
while PowerLawGate and ResonanceCodebook contribute nothing in this tokenized
context.

\subsection{Byte-Level Predecessor Ablation}\label{sec:v7-ablation}

Before arriving at \KathleenClean{}, we developed a larger predecessor
model (1.8M parameters) that additionally incorporated associative memory,
hierarchical key generation, and a bio-inspired gating framework
(``Phantasy''---a multi-stream architecture inspired by cognitive models
of drive, object relations, and memory).
We perform a two-phase ablation on SST-2 to determine which components
justify their parameter cost:

\textbf{Phase~1: Screening} (single seed) identifies relative contributions:

\begin{table}[t]
\centering
\caption{Component ablation of the predecessor model (SST-2, seed=42).
Each row removes one component from the full model.}
\label{tab:ablation-v7}
\begin{tabular}{lccc}
  \toprule
  Variant & SST-2 (\%) & Params & $\Delta$ \\
  \midrule
  Full (all components)     & 84.4 & 1,823K & ---    \\
  $-$ PhaseHarmonics        & 81.8 & 1,364K & $-$2.6 \\
  $-$ HierarchicalKeys      & 82.7 & 1,293K & $-$1.7 \\
  $-$ PhaseShift            & 83.1 & 1,790K & $-$1.3 \\
  $-$ SDM Memory            & 83.4 & 1,823K & $-$1.0 \\
  $-$ Phantasy (560K params) & 84.2 & 1,264K & $-$0.2 \\
  \midrule
  Minimal (none of above)   & 80.4 & 241K   & $-$4.0 \\
  \bottomrule
\end{tabular}
\end{table}

\textbf{Phase~2: Confirmation} (3 seeds) validates:
FULL = 83.6\% $\pm$ 0.6\%, MINIMAL = 81.0\% $\pm$ 0.7\%.

\paragraph{Key findings:}
\begin{enumerate}
  \item \textbf{PhaseHarmonics is MVP}: 6 parameters contribute $+2.6\%$
    ($0.0004$\% of total parameters $\rightarrow$ $65\%$ of total component
    contribution).
  \item \textbf{Phantasy is useless}: 560K parameters (31\% of model)
    contribute only $+0.2\%$---less than the 6-parameter PhaseHarmonics
    by a factor of 13$\times$.
  \item \textbf{Frequency components dominate}: PhaseHarmonics ($+2.6\%$),
    HierarchicalKeys ($+1.7\%$), and PhaseShift ($+1.3\%$) together
    account for $+5.6\%$ of the $+4.0\%$ total gap (components are not
    independent).
\end{enumerate}

This ablation directly informed \KathleenClean{}: we removed Phantasy,
HierarchicalKeys, and SDM (dead weight without Phantasy as consumer),
achieving a 60\% parameter reduction (1.8M $\rightarrow$ 733K) with
minimal accuracy loss.

\subsection{V0 $\to$ V9 Architectural Comparison}\label{sec:v0v9}

Table~\ref{tab:v0v9} isolates the impact of V9's architectural changes
by comparing the same front-end (frequency processing pipeline) with
different back-ends and training protocols:

\begin{table}[t]
\centering
\caption{V0 vs.\ V9 comparison (3-seed averages). V0 uses MLM pretraining;
V9 uses no pretraining.}
\label{tab:v0v9}
\begin{tabular}{lcccccc}
  \toprule
  & \multicolumn{2}{c}{SST-2} & \multicolumn{2}{c}{IMDB} & \multicolumn{2}{c}{AG News} \\
  \cmidrule(lr){2-3}\cmidrule(lr){4-5}\cmidrule(lr){6-7}
  Model & Mean & $\Delta$ & Mean & $\Delta$ & Mean & $\Delta$ \\
  \midrule
  V0 (733K, pretrain) & 83.3\% & --- & 88.6\% & --- & 92.3\% & --- \\
  \textbf{V9 (469K, no pretrain)} & \textbf{85.8\%} & \textbf{+2.5} &
    \textbf{88.5\%} & $-$0.1 & \textbf{92.4\%} & \textbf{+0.1} \\
  \bottomrule
\end{tabular}
\end{table}

The $+2.5\%$ improvement on SST-2 with 36\% fewer parameters and no
pretraining is the most striking result. We attribute this primarily to
Positional Decay Modulation: short texts (${\sim}19$ words $\approx$
80 bytes) have dense, position-sensitive structure where knowing
\emph{where} to retain memory is as important as knowing \emph{what}
to retain. The four-channel architecture with $\Psi$-interference also
contributes by providing explicit pathways for contrast and negation
detection---patterns common in short sentiment texts (``not bad,'' ``hardly
disappointing'').

On IMDB and AG~News, V9 matches V0 within statistical noise, indicating
that these longer-text benchmarks did not require pretraining in the
first place---the V9 architecture is expressive enough to converge to
the same solution from random initialization.

\subsection{PowerLawGate: Context-Dependent Utility}

\begin{table}[t]
\centering
\caption{PowerLawGate effect depends on input representation.}
\label{tab:plg-context}
\begin{tabular}{lc}
  \toprule
  Context & PLG effect \\
  \midrule
  Tokenized (word embeddings)     & 0.0\% \\
  Byte + \texttt{nn.Embedding}    & 0.0\% \\
  \textbf{Byte + FFT-Rotate}      & \textbf{+0.9\%} \\
  \bottomrule
\end{tabular}
\end{table}

This result demonstrates that architectural utility is not intrinsic but
\emph{context-dependent}: the same component can be useless or helpful
depending on its input representation. PLG's power-law compression is
only beneficial when applied to frequency-domain signals with wide
dynamic range, not to bounded embedding outputs.

\subsection{Carrier Cancellation Discovery}\label{sec:carrier}

Early byte-level experiments using sinusoidal carriers
$x(t) = \sin(\omega t + f(\theta_b))$ achieved only 50\% accuracy
(random chance). We diagnosed the root cause: mean pooling destroys the
carrier signal because $\mathbb{E}[\sin(\omega t + \phi)] \approx 0$ for
sufficiently long sequences.

The fix was to remove the carrier oscillation and use only
identity-preserving frequency features (Fourier byte encoding with
$\sin(k\theta)$), immediately recovering 82.3\% accuracy. This
\emph{carrier cancellation} phenomenon may affect other architectures
that combine oscillatory processing with mean pooling.

\section{Architecture Design Process}\label{sec:narrative}

\Kathleen{}'s current architecture emerged through iterative empirical
refinement spanning 20 experimental phases. We describe four pivotal
design decisions.

\paragraph{From generation to classification.}
Initial attempts at autoregressive byte generation with oscillator banks
failed to converge. However, the same components proved effective for
classification, where the task requires \emph{detecting} frequency
patterns rather than \emph{generating} coherent output.

\paragraph{From tokens to bytes.}
Ablation of the tokenized model (Section~\ref{sec:ablation}) revealed
that the ResonanceCodebook---originally the theoretical foundation of
the architecture---contributed $+0.1\%$ in tokenized contexts.
This counter-intuitive finding motivated the shift to raw bytes, where
frequency processing operates on its natural substrate: sequential
signal data rather than discrete symbol embeddings.

\paragraph{Diagnosing carrier cancellation.}
The shift to raw bytes initially produced random-chance accuracy
($50\%$) with sinusoidal carrier approaches. Diagnosing the root cause
(Section~\ref{sec:carrier}) led to Fourier byte encoding ($82.3\%$) and
the FFT-Rotate encoder ($83.7\%$).

\paragraph{From pretraining to architecture.}
Diagnosis of the V0 reverb revealed that its content-dependent decay
$\gamma_t$ remained near-constant ($\gamma \approx 0.94$, $\Delta < 0.5\%$)
throughout training, despite being parameterized to vary. Adding a learnable
positional bias ($+256$ parameters) ``unstuck'' the decay dynamics, enabling
genuinely position-dependent behavior. This single change, combined with
multi-channel processing, eliminated the need for MLM pretraining entirely.

A recurring theme is that \emph{failed experiments yield transferable
components}: FFT-Rotate originated from an unsuccessful language model,
and PowerLawGate proved useful only after the transition to frequency-domain
byte representations.

\section{Discussion}\label{sec:discussion}

\subsection{Strengths}

\paragraph{Pretraining elimination.}
The most significant finding of this work is that architectural design
can substitute for self-supervised pretraining. V9 demonstrates that a
model with the right inductive biases---position-aware decay,
multi-channel interference, energy-proportional mixing---can match a
pretrained model's accuracy while simplifying the training pipeline
from two phases to one.

\paragraph{Extreme parameter efficiency.}
\KathleenV{} achieves $88.5\%$ IMDB and $92.4\%$ AG~News with 469K
parameters---$235\times$ fewer than BERT-base (110M), $281\times$ fewer
than CANINE-S (132M), and $16\times$ fewer than tokenized \Kathleen{}
(11.8M).

\paragraph{No tokenizer.}
\Kathleen{} operates directly on UTF-8 bytes, eliminating:
(1)~language-specific tokenizer training,
(2)~out-of-vocabulary problems,
(3)~tokenization artifacts (subword boundaries obscuring morphology),
(4)~preprocessing pipeline complexity.

\paragraph{$O(L)$ complexity.}
Both time and memory scale linearly, enabling operation at sequence lengths
where $O(L^2)$ Transformers are excluded. This is not merely faster---it
enables fundamentally new use cases (100K+ byte documents, streaming).

\paragraph{Ablation-validated design.}
Every component in \Kathleen{} has been empirically justified through
ablation. The Phantasy framework removal exemplifies principled pruning:
despite theoretical appeal, 560K parameters contributed only $+0.2\%$.

\subsection{Limitations}

\paragraph{Accuracy gap vs.\ pretrained models.}
A ${\sim}7\%$ gap remains vs.\ BERT on SST-2 ($85.8\%$ vs.\ $93.0\%$).
This is primarily a pretraining data gap: BERT uses massive external
corpora (3.3B words from BookCorpus + Wikipedia); \Kathleen{} uses only
task-specific training data. With only 469K parameters, \Kathleen{} cannot
memorize the broad linguistic knowledge that large-scale pretraining provides.

\paragraph{Short-text remains hardest.}
Despite the $+2.5\%$ improvement, SST-2 ($85.8\%$) still lags behind
IMDB ($88.5\%$). Short texts provide fewer bytes for frequency-domain
pattern detection, though Positional Decay Modulation substantially
mitigates this.

\paragraph{Classification only.}
We have not evaluated generation, translation, or other sequence-to-sequence
tasks. The architecture's suitability for autoregressive generation remains
an open question.

\subsection{Future Work}

\begin{itemize}
  \item \textbf{Pretraining on external corpora}: V9 eliminates the need
    for \emph{task-specific} pretraining, but pretraining on a large external
    corpus (like BERT) could close the remaining accuracy gap while
    maintaining the parameter efficiency advantage.
  \item \textbf{Stacked perception layers}: Current V9 uses a single
    processing layer. Stacking 2--4 layers with residual connections could
    close the gap with deeper models.
  \item \textbf{Long-context classification} ($L = 100\text{K}+$): Exploiting
    $O(L)$ complexity for document-level tasks where Transformers cannot
    operate.
  \item \textbf{Edge deployment}: At 469K parameters, \KathleenV{} fits
    on microcontrollers (ESP32) and mobile devices.
  \item \textbf{Streaming classification}: The causal reverb enables
    byte-by-byte processing for real-time applications.
  \item \textbf{Multilingual evaluation}: Byte-level processing is
    inherently language-agnostic; no tokenizer retraining needed.
\end{itemize}

\subsection{Parameter Efficiency Analysis}

Table~\ref{tab:efficiency} compares parameter efficiency across models.

\begin{table}[t]
\centering
\caption{Parameter efficiency comparison (IMDB accuracy).}
\label{tab:efficiency}
\begin{tabular}{lccc}
  \toprule
  Model & IMDB (\%) & Params & Acc/M-params \\
  \midrule
  BERT-base         & 93.0 & 110M   & 0.85 \\
  DistilBERT        & 92.0 & 66M    & 1.39 \\
  CANINE-S          & ---  & 132M   & ---  \\
  \KathleenClean{} (V0) & 88.6 & 733K & 120.9 \\
  \textbf{\KathleenV{}} & \textbf{88.5} & \textbf{469K} & \textbf{188.7} \\
  \bottomrule
\end{tabular}
\end{table}

\KathleenV{} achieves $188.7$ accuracy points per million parameters
on IMDB---$222\times$ more efficient than BERT-base and $56\%$ more
efficient than V0. This extreme efficiency arises from the inductive
bias of frequency processing combined with the multi-channel architecture:
each channel specializes in different aspects of the signal (temporal
memory, local patterns, agreement, disagreement) without parameter
redundancy.

\section{Reproducibility}\label{sec:repro}

All experiments use PyTorch and run on a single NVIDIA T4 GPU (Kaggle
free tier). Training \KathleenV{} takes approximately 30--45 minutes
per dataset per seed. All reported results are mean $\pm$ standard
deviation over seeds $\{42, 123, 456\}$. Datasets are from standard
Hugging Face repositories (\texttt{imdb}, \texttt{ag\_news},
\texttt{glue/sst2}). Code is available at
\url{https://github.com/gfountzoulas/kathleen}.

\section{Conclusion}\label{sec:conclusion}

We presented \Kathleen{}, a frequency-domain architecture for byte-level
text classification that requires no tokenizer, no attention mechanism,
and no self-supervised pretraining. Through systematic ablation and
iterative architectural evolution spanning 20 experimental phases, we
arrived at \KathleenV{}: a 469K-parameter model that processes raw
UTF-8 bytes through frequency-domain feature extraction, multi-channel
processing with consonance and dissonance interference, and
Content-Dependent Reverb with Positional Decay Modulation.

\KathleenV{} achieves $88.5\%$ on IMDB, $92.4\%$ on AG~News, and $85.8\%$
on SST-2---matching or exceeding its pretrained predecessor (V0: $88.6\%$,
$92.3\%$, $83.3\%$) while using 36\% fewer parameters and no pretraining.
The $+2.5\%$ improvement on SST-2 demonstrates that Positional Decay
Modulation specifically addresses the short-text limitation: a single
learnable vector of 256 parameters enables position-aware memory retention
that was previously only achievable through self-supervised pretraining.

\Kathleen{} establishes a new Pareto frontier for efficient byte-level
NLP: $235\times$ fewer parameters than the nearest byte-level competitor
(CANINE-S), with $O(L)$ complexity enabling operation at sequence lengths
where Transformers are fundamentally excluded. The elimination of
pretraining while maintaining accuracy demonstrates that
\emph{architecture is a viable alternative to pretraining} for
resource-efficient text understanding.

\bibliographystyle{plainnat}
\bibliography{references}

\end{document}